\documentclass[lettersize,journal]{IEEEtran}
\usepackage{amsmath,amsfonts}
\usepackage{algorithmic}
\usepackage{algorithm}
\usepackage{array}
\usepackage[caption=false,font=normalsize,labelfont=sf,textfont=sf]{subfig}
\usepackage{textcomp}
\usepackage{stfloats}
\usepackage{url}
\usepackage{verbatim}
\usepackage{graphicx}
\usepackage{cite}
\usepackage{multirow}
\usepackage{hyperref}
\usepackage{booktabs}
\usepackage{tikz}
\usepackage{etoolbox}
\makeatletter
\patchcmd{\@makecaption}
  {\scshape}
  {}
  {}
  {}
\makeatother

\hypersetup{
  colorlinks=true,
  citecolor=blue,
  linkcolor=red,
  urlcolor=blue
  }

\DeclareRobustCommand{\orcidicon}{%
	\begin{tikzpicture}
	\draw[lime, fill=lime] (0,0) 
	circle [radius=0.134] 
	node[white] {{\fontfamily{qag}\selectfont \tiny ID}};    \draw[white, fill=white] (-0.0625,0.095) 
	circle [radius=0.007];    \end{tikzpicture}
	\hspace{-2mm}}
\foreach \x in {A, ..., Z}{%
	\expandafter\xdef\csname orcid\x\endcsname{\noexpand\href{https://orcid.org/\csname orcidauthor\x\endcsname}{\noexpand\orcidicon}}
}

\begin{document}

\newcommand{\orcidauthorA}{0000-0002-7695-1633} %ZYF
\newcommand{\orcidauthorB}{0009-0003-8889-8985} %LC
\newcommand{\orcidauthorC}{0009-0009-3270-1526} %YHJ
\newcommand{\orcidauthorD}{0000-0001-5571-8018} %FT
\newcommand{\orcidauthorE}{0000-0002-9274-0801} %ZZX
\newcommand{\orcidauthorF}{0000-0003-0663-332X} %ZPF
\newcommand{\orcidauthorG}{0000-0002-6769-3177} %OZH

\title{5\% $>$ 100\%: Flatness Preference is All You Need for Multimodal Parameter-Efficient Fine-Tuning}

\author{Yifan~Zhu\orcidA{},~\IEEEmembership{Senior Member,~IEEE,}
        Can~Lin~\orcidB{},
        Hangjie~Yuan~\orcidC{},
        Zixiang~Zhao~\orcidE{},
        Pengfei~Zhang~\orcidF{},
        Tao~Feng~\orcidD{},
        Zhonghong~Ou~\orcidG{}% <-this % stops a space

\thanks{This work is supported by the National Key Research and Development Program of China under Grant 2024YFC3308500, Beijing Municipal Natural Science Foundation under Grant L251042, National Natural Science Foundation of China under Grant 62406036, and also sponsored by the State Key Laboratory of Networking and Switching Technology under Grant NST20250110. (Corresponding author: Tao Feng.)
}% <-this % stops a space
\thanks{Yifan Zhu and Can Lin are with the School of Computer Sciences, Beijing University of Posts and Telecommunications, Beijing 100876, China (e-mail: yifan\_zhu@bupt.edu.cn; lincan@bupt.edu.cn).}
\thanks{Hangjie Yuan is with the College of Computer Science and Technology, Zhejiang University, Zhejiang 310027, China (e-mail: hj.yuan@zju.edu.cn).}%
\thanks{Zixiang~Zhao is with ETH Zürich, Switzerland (e-mail: zixiang.zhao@hotmail.com).}
\thanks{Pengfei~Zhang is with Anhui University of Science and Technology, Anhui 232001, China (e-mail: zpf.bupt@bupt.cn).}%
\thanks{Tao Feng is with the Department of Computer Science and Technology, Tsinghua University, Beijing 100084, China (e-mail: fengtao.hi@gmail.com).}%
\thanks{Zhonghong Ou is with State Key Laboratory of Networking and Switching Technology, Beijing University of Posts and Telecommunications, Beijing 100876, China, and also with the School of Computer Sciences, Beijing University of Posts and Telecommunications, Beijing, China (e-mail: zhonghong.ou@bupt.edu.cn).} }%

\markboth{IEEE Transactions on Multimedia,~Vol.~XX, No.~XX, XX~2026}%
{Shell \MakeLowercase{\textit{et al.}}: A Sample Article Using IEEEtran.cls for IEEE Journals}


\maketitle

\begin{abstract}
Parameter-Efficient Fine-Tuning (PEFT) methods provide a streamlined and efficient tool for adapting large models to domain-specific multimodal downstream tasks. Although these methods proved their tangible effects in practice, their principal aspects remain under-explored. Therefore we remain curious about the underlying generalization mechanisms in various PEFT methods and how they can be further enhanced. In this paper, we reveal the flatness preference widely present in various PEFTs, where a small fraction of sharp dimensions dominates the generalization of PEFT. This finding suggests an appealing possibility: we may be satisfied with a better generalization by merely attending to this small fraction of sharp dimensions instead of all of them. Furthermore, we propose \textbf{Flat}ness \textbf{P}reference \textbf{O}ptimization (FlatPO) to flatten these key sharpness dimensions, leading various PEFTs toward better generalization. Extensive experiments demonstrate the effectiveness of our findings and the proposed method. Code is available at \url{https://github.com/Can-Lin/FlatPO}. 
\end{abstract}


\section{Introduction}
\IEEEPARstart{P}{arameter}-Efficient Fine-Tuning (PEFT) offers a promising solution for adapting large models to various multimodal downstream tasks~\cite{han2024parameter, wang2025parameter}. This technique focuses on updating a few parameters to achieve comparable performance to full fine-tuning at a significantly low cost~\cite{xu2026parameter}. Common methods include Low-Rank Adaptation (LoRA)~\cite{hu2021lora}, prompt tuning~\cite{lester2021power}, adapter tuning~\cite{hu2023llm}, \emph{etc}. 
These methods encompass a range of fine-tuning strategies~\cite{han2024parameter}, such as reparameterization and additive adaptation and have also been actively explored in multimedia scenarios~\cite{10171397,zhang2024unleash,sun2026unleashing}.

Although different PEFTs vary in their trainable sets (\emph{i.e.}, the structure, modules, and position of trainable parameters)~\cite{han2024parameter}, they fundamentally share the principle of fine-tuning a small fraction of parameters to steer the pre-trained model loss toward a relatively optimal minimum in the downstream domain. Thus, the success of PEFTs can be summarized as a result of generalization~\cite{oikonomou2025sharpness,wang2025parameter,liu2025bi}. 
More specifically, pre-training initialization offers a favorable starting point, reflected as a prominent initial position in the loss landscape~\cite{song2023sparse,chen2025understanding}. Subsequently, the optimizer focuses on reducing the loss to guide the model toward improved generalization~\cite{zhou2024towards,deng2025eflat}, which reveals the significance of using PEFT tools. However, current PEFTs do not explicitly incorporate generalization as a guiding principle during optimization. This often causes the optimization to converge to sharp minima, undermining downstream generalization performance.

\begin{figure}[t]
    \centering
    \includegraphics[width=\linewidth]{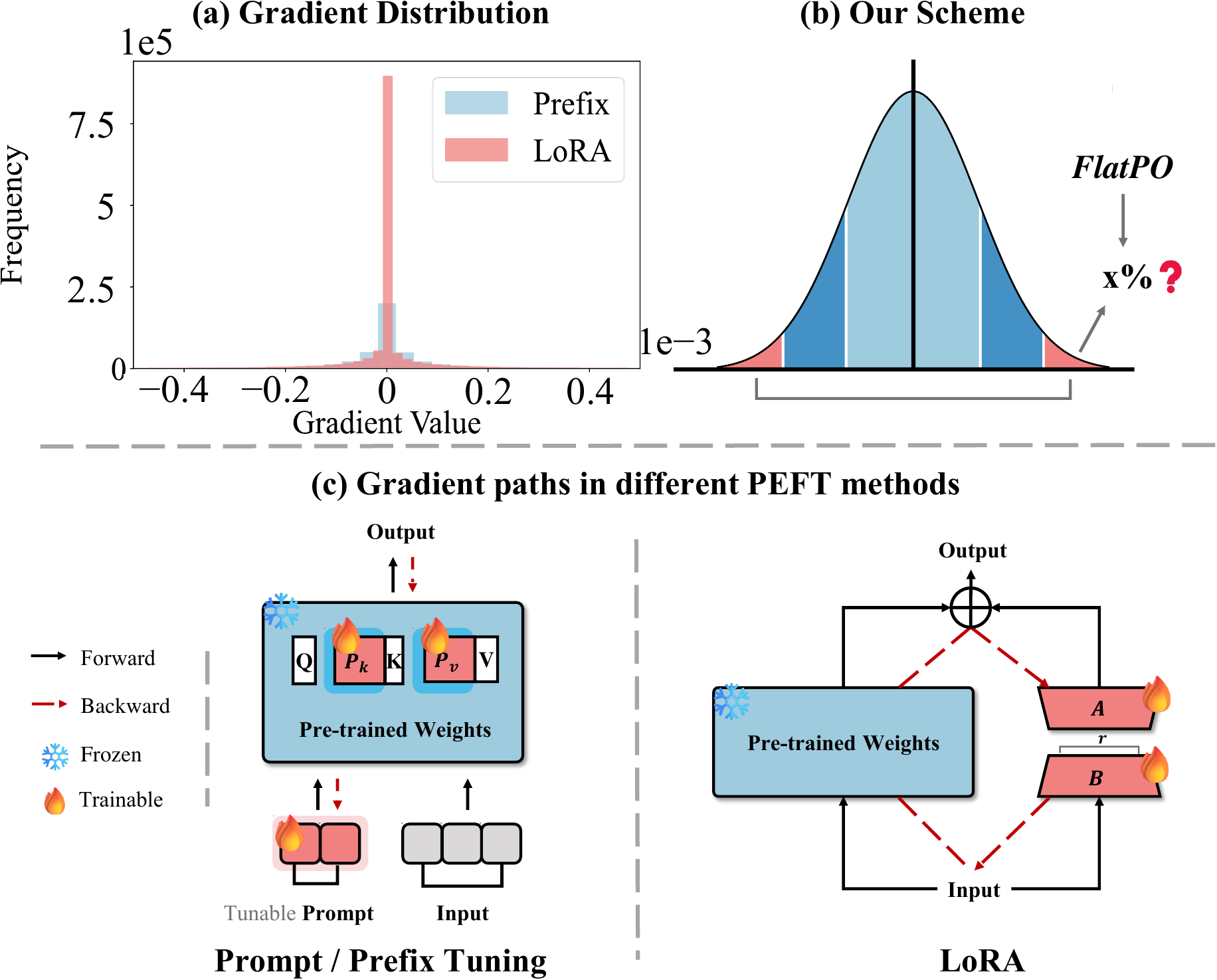}
    \caption{Sharp dimensions in parameter gradients are the critical minority affecting fine-tuning generalization (a) The parameter gradient distributions (total) on LoRA and prefix tuning, indicating flat dimensions dominate the loss landscape while sharp dimensions occupy a minor portion. (b) Our FlatPO assumes that focusing on the top \textbf{$x\%$} of sharp values may lead to better generalization, as these high-magnitude gradients typically contribute the most significant updates. (c) Various PEFTs feature different backward gradient flows (red dotted lines), which implies the diversity and randomness of flat solutions in PEFTs.}
    \label{fig:intro}
    %\vspace{-15pt}
\end{figure}

A new trend highlights that flat solution bootstraps better generalization in fine-tuning~\cite{deng2025eflat,song2023sparse,li2024flat}. Specifically, Pre-trained models can easily identify flat minima, ensuring robust generalization even with a few trainable parameters~\cite{song2023sparse,li2024flat}. They reveal a pattern of sparsity in parameter gradients for downstream tasks, where 1\% of the components dominate 99\% of the gradient norm~\cite{song2023sparse,wang2024improving}. As Figure~\ref{fig:intro}(a) shows, this finding also appears in PEFT. It indicates that flat dimensions dominate the loss landscape during fine-tuning, while sharp dimensions occupy a minor portion~\cite{song2023sparse,lee2023domain}. Consequently, uniformly flattening all dimensions for fine-tuning may be unnecessary and may not even be optimal, which is transferred to two key insights: \textit{(i) the flat solution is equally important in fine-tuning; (ii) the flat solution may exhibit a preference in fine-tuning.} Therefore, we only need to focus on a few sharp dimensions, as shown in Figure~\ref{fig:intro}(b), which is enough to derive better generalization.

Sharp values still plague various PEFT methods~\cite{chen2023ptp}, particularly within local regions of the low-dimensional optimization space~\cite{xu2026parameter,li2024flat}. In particular, due to the updates to the limited parameters, the PEFT optimization trajectories are typically short~\cite{saha2026grit}. This limited search space often results in solutions concentrated within local regions that are prone to sharp minima~\cite{lee2023domain}. Furthermore, the diversity of PEFT strategies amplifies the randomness of sharp value occurrences. The differences in trainable sets across various PEFTs make the position of sharp values unpredictable~\cite{lee2023domain,chen2023ptp} and Figure~\ref{fig:intro}(c) illustrates this diversity through the distinct backward gradient flows in different PEFTs, which means that uniformly flattening a particular layer or component across PEFTs is a challenging task. Although general flatness optimization methods, like Sharpness-Aware Minimization (SAM)~\cite{zhou2025sharpness}, aim to achieve flat solutions, they merely unlock the potential for flat solutions in PEFT without accommodating the diverse optimization demands of different PEFTs. For instance, these methods struggle to adapt to varying flatness preferences and the distinct paradigms of PEFT.

To this end, in this paper, we propose \textbf{Flat}ness \textbf{P}reference \textbf{O}ptimization (FlatPO) to enhance the generalization capabilities of various PEFTs. First, we evaluate the gradient distribution of the trainable parameters, thereby obtaining their flatness preferences. Next, we prioritize optimization based on the impact of each dimension on the loss landscape, selectively smoothing sharp dimensions while retaining the parts that are vital for generalization. Then, FlatPO is seamlessly integrated into various PEFTs to satisfy their unique flatness requirements. Overall, FlatPO places a preference on the 5\% sharpness dimensions and offers solid gains.

To summarize, our contributions are listed below:
\begin{itemize}
\item We reveal a prevalent flatness preference across various PEFTs, \emph{i.e.}, 5\% of the gradient parameters more urgently desire flatness for seeking minima.
\item We propose a flatness preference optimization (FlatPO) that senses and flattens the dominant sharpness in PEFTs to efficiently find flat minima.
\item Extensive results prove the versatility of FlatPO, which guides various PEFTs to reach better generalization. This paves the way for more streamlined multimodal fine-tuning.
\end{itemize}

\par The remainder of this paper is organized as follows. Section \ref{Section2} reviews the related work on PEFTs and flat minima. Section \ref{section3} presents the preliminaries of our study, including the problem definition and the SAM strategy. Section \ref{section4} analyzes the flatness preference phenomenon in PEFT and introduces the proposed FlatPO method in detail. Section \ref{section5} provides extensive experimental results, ablation studies, and discussions. Finally, Section \ref{section6} concludes our work.

\section{Related Work}
\label{Section2}

In this section, we first review recent advances in PEFT. Then, we introduce representative studies on flat minima and generalization in fine-tuning.
% \subsection{Parameter Efficient Fine Tuning}
% In general, PEFT methods are typically categorized into four main categories: selective PEFT, additive PEFT, reparameterized PEFT, and hybrid one~\cite{han2024parameter}. Among them, additive PEFT and reparameterized PEFT are the two prevalent groups. In the former, adapter-based and prompt-based methods are more commonly used. This type of method introduces a few new tunable parameters within transformer module or input layer to perform fine-tuning, such as Serial Adapter~\cite{houlsby2019parameter}, AdaMix~\cite{wang2022adamix}, Contrl Adapter~\cite{lin2024ctrl}, Adapter-X~\cite{li2024adapter}, Prompt Tuning~\cite{lester2021power}, Prefix-Tuning~\cite{li2021prefix}, Visual Prompt Tuning~\cite{jia2022visual}, and CP-Prompt Tuning~\cite{CP-Prompt}, \emph{etc}. 
% In the latter, a line of work along the direction of LoRA, \emph{e.g.}, LoRA~\cite{hu2021lora}, AdaLoRA~\cite{zhang2023adalora}, LoRA+~\cite{hayou2024lora+}, MOELoRA~\cite{liu2023moelora}, MoLE~\cite{chen2024llava}, \emph{etc}.{These methods span various models and applications, including Large Language Model(LLM), Vision-Language Model(VLM), Multimodal Large Language Model(MLLM), language or image understanding, or text-to-image models.

\subsection{Parameter Efficient Fine-Tuning}
In general, PEFT methods are typically categorized into four main categories: selective PEFT, additive PEFT, reparameterized PEFT, and hybrid one~\cite{han2024parameter}. Among them, additive PEFT and reparameterized PEFT are the two prevalent groups. Additive PEFT introduces a small number of additional trainable parameters into the transformer blocks or input layer while freezing the backbone parameters. In this category, adapter-based and prompt-based methods are the most commonly used. Representative approaches include Serial Adapter~\cite{houlsby2019parameter}, AdaMix~\cite{wang2022adamix}, Ctrl-Adapter~\cite{lin2024ctrl}, Adapter-X~\cite{li2024adapter}, Prompt Tuning~\cite{lester2021power}, Prefix-Tuning~\cite{li2021prefix}, Visual Prompt Tuning~\cite{jia2022visual}, and CP-Prompt Tuning~\cite{CP-Prompt}, \emph{etc}. Reparameterized PEFT typically reformulates parameter updates into low-rank or other structured forms. A representative line of work in this direction is based on LoRA, including LoRA~\cite{hu2021lora}, AdaLoRA~\cite{zhang2023adalora}, LoRA+~\cite{hayou2024lora+}, MOELoRA~\cite{liu2023moelora}, and MoLE~\cite{chen2024llava}, \emph{etc}. These methods have been applied to a wide range of models and tasks, including LLMs, VLMs, MLLMs, language and image understanding, as well as text-to-image generation.

\subsection{Flat Minima and Generalization in Fine-tuning}
Advanced generalization improvements are vital for fine-tuning techniques. PromptSRC~\cite{khattak2023self} and CoPrompt~\cite{roy2023consistency} mitigate overfitting by preserving the knowledge of the pre-trained model or enforcing prediction consistency. C-Flat~\cite{bian2024make} promotes better generalization by shaping a flatter loss landscape. PACE~\cite{ni2024pace} connects smaller gradient norms with improved generalization in PEFT and introduces consistency regularization to align fine-tuned models with their pre-trained counterparts. GLAD~\cite{peng2025glad} enhances generalization by applying gradient regularization to LoRA-based tuning. These efforts highlight the need to focus on generalization in fine-tuning.

Another line of work aims to improve generalization by explicitly seeking flat minima. Sharpness-Aware Minimization (SAM)~\cite{foret2020sharpness} provides a general framework by minimizing the worst-case loss in a local neighborhood, while Gradient norm Aware Minimization (GAM)~\cite{zhang2023gradient} further characterizes flatness through the maximum gradient norm and connects first-order flatness with the Hessian spectrum. Recent studies extend these ideas to LoRA. Flat-LoRA~\cite{li2024flat} points out that flatness in the LoRA optimization space does not necessarily imply flatness in the original full-parameter space, and therefore seeks low-rank adaptations located in flat regions of the full parameter landscape. EFlat-LoRA~\cite{deng2025eflat} improves the efficiency of this idea and empirically verifies the connection between sharpness and LoRA generalization on both LLMs and VLMs. Bi-LoRA~\cite{liu2025bi} revisits SAM under LoRA fine-tuning and introduces an auxiliary adversarial LoRA branch to model broader sharpness while avoiding the doubled cost of standard SAM. Currently, attention to flat minima in PEFT is still in its infancy~\cite{song2023sparse,ding2023parameter}. Nevertheless, when generalizing flat minima to PEFTs, we must be mindful of the relationship between the fine-tuning optimization space and the original full-parameter space across various PEFTs.

\section{Preliminaries}
\label{section3}

In this section, we introduce the preliminary knowledge of our study. We begin with the problem definition of PEFTs, and then present the SAM strategy that motivates our method.

\subsection{Problem Definition}
Parameter-Efficient Fine-Tuning (PEFT) refers to a family of methods designed to adapt pre-trained models to downstream tasks with minimal updates to the model parameters. Given a pre-trained model \( M \), PEFT aims to adapt it to a downstream task \( T \) with minimal modifications to the model parameters. Let the original model parameters be \( W_0 \), and the input to the model be \( x_{\text{in}} \). The goal of PEFT is to find a set of task-specific parameters \( W_T \), where \( W_T \subset W_0 \) or is an auxiliary set of parameters that are much smaller than \( W_0 \), \emph{i.e.}, $|W_T| \ll |W_0|$ while keeping the performance in the downstream task \( T \) is maximized. In our framework, we examine three widely used methods for PEFT: \textbf{LoRA}~\cite{hu2021lora}, \textbf{Prefix Tuning}~\cite{li2021prefix}, and \textbf{Prompt Tuning}~\cite{lester2021power} to evaluate the effectiveness of our approach. In LoRA,\( W_T \) corresponds to the low-rank matrices \( W_{\text{up}} \) and \( W_{\text{down}} \). In Prefix / Prompt Tuning, \( W_T \) refers to the learned embeddings \( P \).

\textbf{LoRA.} LoRA introduces low-rank trainable matrices \( W_{\text{up}} \) and \( W_{\text{down}} \) to pre-trained weights \( W_0 \), keeping the original weights frozen. The output with LoRA is denoted as:
\begin{equation}
x_{\text{out}} = W_0 x_{\text{in}} + \frac{\alpha}{r} W_{\text{up}} W_{\text{down}} x_{\text{in}},
\end{equation}
where \( \alpha \) is a scaling factor and \( r \) is the rank of the matrices.

\textbf{Prefix Tuning.} Prefix Tuning adds task-specific prefix embeddings \( P \) to the input sequence \( x_{\text{in}} \). The input becomes:
\begin{equation}
x_{\text{aug}} = [P; x_{\text{in}}],
\end{equation}
where \( P \in \mathbb{R}^{L_{\text{prefix}} \times d} \) and \( x_{\text{in}} \) is the original input.

\textbf{Prompt Tuning.} The prompt consists of continuous embedding vectors \( P \in \mathbb{R}^{L_{\text{prompt}} \times d} \). The prompt is prepended to the original input sequence \( x_{\text{in}} \) and passed through the model.

It is important to note that Prefix Tuning and Prompt Tuning differ significantly in their implementation. The former adds learnable continuous vectors before the key/value sequences in every layer of the Transformer, while the latter only incorporates tunable prompt vectors at the input layer.

\subsection{Sharpness-Aware Minimization (SAM)}

Instead of simply minimizing the training loss \(L_S\), SAM \cite{foret2020sharpness} aims to guide the training process towards convergence in a flatter region. To achieve this, SAM introduces a training scheme that solves the following Min-Max Optimization Problem:
\begin{equation}
\min_{w} \max_{\|\varepsilon\|_2 \leq \rho} L_S(f_{w + \varepsilon}),
\end{equation}
where \( \rho \) is a predefined neighborhood radius, and \( \varepsilon \) is the weight perturbation vector that maximizes the training loss within the \( \rho \)-constrained neighborhood. 

The idea of applying a uniform perturbation to all parameters does not seem entirely consistent with the concept of PEFT. PEFT emphasizes that adjusting only a small subset of key parameters can achieve effective fine-tuning while significantly reducing computational and storage costs. In the context of PEFT, many parameters are actually located in a relatively flat minimum and do not require additional perturbation. Building on this insight, we can make a targeted modification to the SAM by applying perturbations solely to those parameters most relevant to the fine-tuning objective. This approach not only aligns more closely with the core principles of PEFT, but also maintains efficiency and further enhances the generalization performance of the model.

\section{Understanding Flatness Preference in PEFT}
\label{section4}

\begin{figure*}[t]
    \centering
    \includegraphics[width=\linewidth]{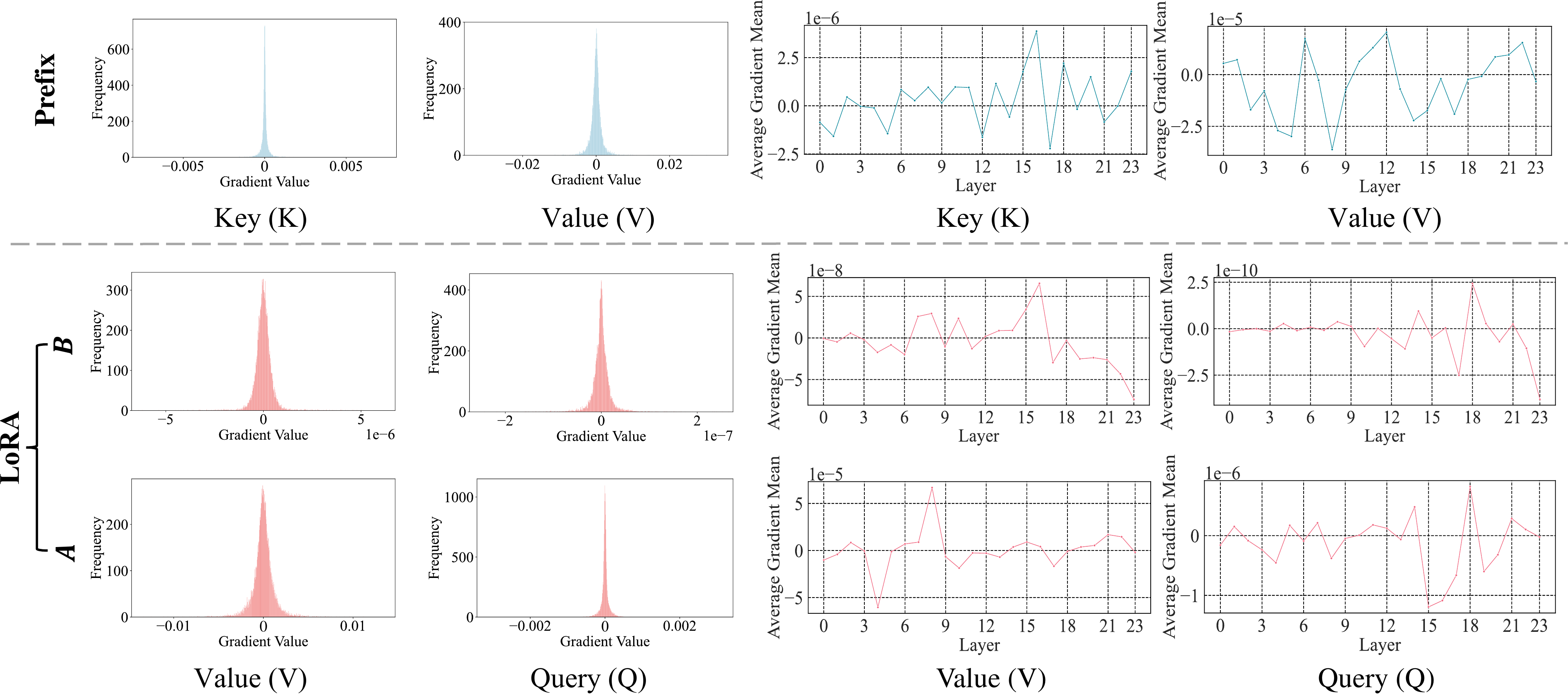}
    %\vspace{-6mm}
    \caption{\textbf{Left} shows the gradient distribution of the parameters in the same layer for LoRA and prefix tuning. In LoRA, $A$ is the down-projection matrix and $B$ is the up-projection matrix. The gradient values for each method in this layer are primarily concentrated around zero, indicating the presence of many flat dimensions in the loss landscape. \textbf{Right} illustrates the variation trends of the average gradient across different layers. Between layers, the mean gradients exhibit varying degrees of random fluctuations, reflecting the differences in optimization paths for different PEFT methods.}
    \label{fig:understand}
    %\vspace{-4mm}
\end{figure*}

% Here, we employ the OPT-1.3B~\cite{zhang2022opt} to analyze two typical PEFT methods (LoRA and Prefix Tuning) on SST2~\cite{wang2019superglue}. 
In this section, we focus on the characteristics of gradient distributions and the trends in optimization paths, and then introduce the formulation of flatness preference together with the proposed optimization strategy.

\subsection{A Thorough Exploration of Gradient Distribution}

As shown in Figure~\ref{fig:understand} (left), we plot the distribution of the parameters’ gradients from LoRA and Prefix Tuning at the same layer, with emphasis on the parameter matrices that they respectively affect, \textit{Query (Q)} and \textit{Value (V)} from LoRA, \textit{Key (K)} and \textit{Value (V)} from Prefix Tuning.

Overall, the parameter gradients of both LoRA and prefix tuning show a clear sparse pattern: most gradient values are concentrated around 0, with only a small fraction having larger magnitudes. LoRA introduces trainable low-rank matrices in the \textit{Q} and \textit{V} parts of the attention layer. Consequently, LoRA's gradients tend to be more concentrated within certain intervals. This indicates that LoRA tends to apply more significant updates in a few dimensions. Prefix tuning guides the model to focus on specific task-related information during the context construction phase of the attention layer by inserting prefix vectors, primarily in the \textit{K} and \textit{V} parts. Unlike LoRA, the gradient distribution of Prefix tuning is more diffuse, with moderately large gradient values scattered more randomly across different dimensions. 

In summary, both PEFTs show that flat dimensions dominate the gradient norm during fine-tuning, while sharp dimensions occupy a minor portion. This observation makes a strong statement about our motivation about gradient preferences.

\subsection{Diversity of the Optimization Path in PEFTs}
Various PEFTs feature different optimization paths, which trigger different sparsity patterns. Here, we draw plots of how the mean of each layer's gradients varies with the layer increases, revealing the dynamic changes of gradients values in the optimization paths of LoRA and Prefix Tuning across all 24 layers of the transformer. 

As shown in Figure~\ref{fig:understand} (right), as the layer increases, the mean gradients of \textit{Q} and \textit{V} in LoRA are generally small and relatively smooth. However, in certain layers closer to the output, the mean gradients of \textit{Q} and \textit{V} exhibit more pronounced fluctuations, corroborating the concentrated update characteristic observed in the histograms. In contrast to LoRA's concentrated fluctuations, prefix tuning shows a more random fluctuation of mean gradients in \textit{K} and \textit{V} across the layers, without any notable surge near the output layers. In other words, Prefix Tuning updates \textit{K} and \textit{V} more randomly across different layers, and any layer could experience a sizable gradient update. Nevertheless, the magnitude of these updates is generally less extreme than LoRA's specific spikes.

Overall, this observation exemplifies the diversity of the sparse patterns resulting from different optimization paths in various PEFTs, making a strong statement about another motivation we mentioned above.

\subsection{Inherent Flatness Preference}
In previous analysis, we identified a general trend in various PEFTs, \emph{i.e.}, flat dimensions dominate the gradient norm, while only a tiny fraction of parameter dimensions exhibit sharpness. Here, we hypothesize that these sharp dimensions constitute \textbf{\(\alpha\%\)} of all dimensions. In detail, this trend means that most flat dimensions remain stable during fine-tuning, and these dimensions make finding a flat minimum less difficult. In contrast, other sharp dimensions (\textbf{\(\alpha\%\)}) are assumed to be crucial to enhancing the generalization capabilities.

\subsection{Rethinking the Gradient Contribution}
To quantify how much optimization-relevant information is preserved under different sparsity levels (\(\alpha\%\)), we analyze gradient contribution from an information-theoretic perspective. Following the observation that gradient distributions in PEFT exhibit similar quasi-sparse properties~\cite{song2023sparse}, we use the 3\(\sigma\) rule only as a heuristic initializer for identifying statistically significant gradient outliers, rather than as a probabilistic guarantee under an exact Gaussian assumption. In practice, we observe that approximately the top \(5\%\) of parameters often correspond to gradient values exceeding roughly \(2\sigma\) away from the mean, indicating that a small subset of dimensions may dominate the optimization dynamics. The final choice of \(\alpha\%\) is further validated by the gradient contribution curve (Figure~\ref{fig:Gradient}) and the performance ablations (Table~\ref{tab:rate}).

We refer to this phenomenon as \textbf{Flatness Preference}, \emph{i.e.}, a small subset of sensitive dimensions exhibits a stronger need for flatness-oriented optimization in order to facilitate the search for flatter minima. Moreover, our preliminary study shows that different PEFT methods follow diverse optimization paths, which implies that simply flattening specific layers or modules is insufficient. This diversity also makes the exact value of \(\alpha\%\) task- and method-dependent.

\subsection{Quantitative Measurement of Flatness Preference}
% In our framework, the notion of \textbf{Flatness Preference} is designed to capture how ``flat" the local region around a particular parameter is in the loss landscape. 
% Intuitively, ``flatness" reflects how sensitive the loss function is to small perturbations of model parameters: if a slight variation in a parameter does not significantly increase the loss, we consider the local region around that parameter to be relatively flat; otherwise, it is sharp. To quantify this \textbf{flatness preference}, we need a practical way to assess such flatness in a local neighborhood of the current parameters.
In our framework, the notion of \textbf{Flatness Preference} is intended to characterize how ``flat'' the local region around a parameter is in the loss landscape. Intuitively, flatness reflects how sensitive the loss is to small perturbations of model parameters: if a small change in a parameter does not significantly increase the loss, the corresponding local region is regarded as flat; otherwise, it is sharp. To quantify this \textbf{flatness preference}, we need a practical way to assess such flatness in a local neighborhood of the current parameters.

The most rigorous characterization of local curvature emerges from the spectral properties of the Hessian matrix:
\begin{equation}
  H_{t} = \nabla^2 L_{\text{oracle}}(\theta_t) \;\in\; \mathbb{R}^{d \times d}.
\end{equation}
The diagonal entries of the matrix can be formulated as:
\begin{equation}
  H_{t}^{\mathrm{diag}}[i] 
  \;=\; 
  \frac{\partial^2 L_{\text{oracle}}(\theta_t)}{\partial \theta_{t,i}^2}
\end{equation}
which represents the curvature along each parameter dimension. In practice, computing per-parameter curvature (e.g., Hessian diagonal) is expensive for large-scale models.
%However, the computational complexity of Hessian estimation scales quadratically with parameter dimensionality, rendering this approach intractable for contemporary large-scale neural architectures.

In practice, first-order methods often provide a good balance between computational tractability and flatness estimation. If a small perturbation to parameter \(\theta_{t,i}\) significantly changes the loss \(L_{\text{oracle}}(\theta_t)\), that parameter lies in a relatively sharp region. \(\theta\) refers to the PEFT parameters that are to be optimized, while the original LLM parameters remain frozen. To measure this preference, we rely on a first-order sensitivity proxy.
%To measure this preference, we rely on the magnitude of the gradient.
%A large value implies a less flatness, while a smaller value indicates a flatter neighborhood.
A larger gradient magnitude indicates higher local sensitivity to small perturbations (i.e., a sharper direction), while a smaller value suggests a flatter neighborhood. We denote this gradient-based measure as the \textbf{Flatness Preference}:
\begin{equation}
      S_{\mathrm{grad}}(\theta_{t,i}) = \left| \frac{\partial L_{\text{oracle}}(\theta_t)}{\partial \theta_{t,i}} \right|
\end{equation}

where \(\theta_t\) is the current parameter vector at iteration \(t\), \(L_{\mathrm{oracle}}\) refers to the standard cross-entropy loss.
Throughout this work, we adopt \( S_{\mathrm{grad}}(\theta_{t,i}) \) as our primary metric for quantifying flatness preference in the parameter space.

\begin{algorithm}[t]
\caption{Flatness Preference Optimization (FlatPO)}
\label{alg:fpo}
\begin{algorithmic}[1]
\STATE \textbf{Input:} Training data, batch size \(b\), learning rate \(\eta_t\), perturbation radius \(\rho_t\), selection ratio \(\alpha\%\)
\STATE \textbf{Initialize:} \(\theta_0\), iteration \(t \gets 0\)
\WHILE{\emph{not converged}}
   \STATE Sample mini-batch \(W_t\) of size \(b\)
   \STATE Compute gradient: \(g_t \gets \nabla L_{\mathrm{oracle}}(\theta_t)\)
   \STATE Compute sensitivity scores: \(\mathcal{M}[i] \gets |g_t[i]|\)
   \STATE Select top-\(\alpha\%\) indices:
   \[
      I_{\mathrm{selected}}
      \gets
      \operatorname{TopIndex}_{\alpha\%}
      \bigl(\{\mathcal{M}[i]\}_{i=1}^{d}\bigr)
   \]
   \STATE Construct sparse perturbation:
   \[
      \Delta \theta_t
      \gets
      \rho_t \,
      \frac{
        g_t \odot \mathbb{I}_{I_{\mathrm{selected}}}
      }{
        \|g_t \odot \mathbb{I}_{I_{\mathrm{selected}}}\|_2
      }
   \]
   \STATE \textbf{if} \(\|g_t \odot \mathbb{I}_{I_{\mathrm{selected}}}\|_2 = 0\) \textbf{then} \(\Delta \theta_t \gets 0\)
   \STATE Perturb parameters: \(\theta_t^{\mathrm{perturbed}} \gets \theta_t + \Delta \theta_t\)
   \STATE Compute perturbed gradient:
   \[
      g_t^{\mathrm{perturbed}}
      \gets
      \nabla L_{\mathrm{oracle}}(\theta_t^{\mathrm{perturbed}})
   \]
   \STATE Update parameters:
   \[
      \theta_{t+1}
      \gets
      \theta_t - \eta_t \cdot g_t^{\mathrm{perturbed}}
   \]
   \STATE \(t \gets t + 1\)
\ENDWHILE
\STATE \textbf{Return:} \(\theta_t\)
\end{algorithmic}
\end{algorithm}

\subsection{Flatness Preference Optimization for PEFT}
We extend the PAC-Bayesian generalization bound analysis to PEFT scenarios. The key insight is that pre-training shifts the prior distribution, leading to quasi-sparse gradient patterns where effective fine-tuning can be achieved by updating only a subset of critical parameters. Consequently, after estimating the flatness preference of each parameter, we focus on a small subset of critical parameters for the targeted perturbation and update. This procedure is designed to align naturally with PEFT, thereby preserving their parameter-efficiency characteristics while encouraging flatter minima. 
At iteration \(t\), let
\begin{equation}
  g_t = \nabla L_{\mathrm{oracle}}(\theta_t)
\end{equation}
denote the gradient with respect to the trainable PEFT parameters. We first define a sensitivity vector \(\mathcal{M}\) whose \(i\)-th entry is the gradient magnitude:
\begin{equation}
  \mathcal{M}[i]
  =
  S_{\mathrm{grad}}(\theta_{t,i})
  =
  |g_t[i]|.
\end{equation}
We then rank these scores in descending order and select the indices corresponding to the top \(\alpha\%\) entries:
\begin{equation}
  I_{\mathrm{selected}}
  =
  \operatorname{TopIndex}_{\alpha\%}
  \Bigl(
    \{\mathcal{M}[i]\}_{i=1}^{d}
  \Bigr).
\end{equation}
By focusing on the top \(\alpha\%\) of parameters with the highest sensitivity, we follow the sparsity phenomenon observed in PEFT, where most parameters remain unchanged or only mildly updated.

Let \(\mathbb{I}_{I_{\mathrm{selected}}} \in \{0,1\}^d\) denote the binary mask corresponding to the selected indices. Inspired by SAM, we construct a perturbation only on the selected sensitive dimensions:
\begin{equation}
  \Delta \theta_t
  =
  \rho_t \,
  \frac{
    g_t \odot \mathbb{I}_{I_{\mathrm{selected}}}
  }{
    \left\|
      g_t \odot \mathbb{I}_{I_{\mathrm{selected}}}
    \right\|_2
  },
\end{equation}
where \(\rho_t\) is the perturbation radius and \(\odot\) denotes element-wise multiplication. When \(\|g_t \odot \mathbb{I}_{I_{\mathrm{selected}}}\|_2 = 0\), we simply set \(\Delta \theta_t = 0\).

We then perturb the current parameters:
\begin{equation}
  \theta_t^{\mathrm{perturbed}}
  =
  \theta_t + \Delta \theta_t,
\end{equation}
and compute the gradient at the perturbed point:
\begin{equation}
  g_t^{\mathrm{perturbed}}
  =
  \nabla L_{\mathrm{oracle}}
  \bigl(
    \theta_t^{\mathrm{perturbed}}
  \bigr).
\end{equation}
Finally, the PEFT parameters are updated by
\begin{equation}
  \theta_{t+1}
  =
  \theta_t - \eta_t g_t^{\mathrm{perturbed}},
\end{equation}
where \(\eta_t\) is the learning rate.

Since the perturbation is applied only to a small subset of sensitive parameters, the additional training overhead can be reduced compared with perturbing all trainable parameters indiscriminately. Meanwhile, the proposed FlatPO framework is plug-and-play for different PEFT methods, with the goal of stabilizing training dynamics and improving generalization performance.

We summarize the pseudo code of FlatPO in Algorithm~\ref{alg:fpo}.

\begin{table*}[b]
\centering
\setlength{\tabcolsep}{6pt}
\captionsetup{font={scriptsize}, labelsep = newline, justification=centering}
\caption{Fine-tuning performance of \textbf{LLM} with typical PEFT methods (LoRA, Prefix Tuning, and Prompt Tuning) on the GLUE (SST2) and SuperGLUE (WIC, WSC and BoolQ) benchmarks. The best results are highlighted in \textbf{bold}.}
\label{tab:main}

\newcolumntype{C}[1]{>{\centering\arraybackslash}m{#1}}

\begin{tabular}{C{1.8cm} C{1.6cm} *{8}{C{0.95cm}}}
\toprule[1.2pt]

\multirow{2}{*}{\textbf{MODEL}} & \multirow{2}{*}{\textbf{METHOD}}
& \multicolumn{4}{c}{\textbf{SST2}}
& \multicolumn{4}{c}{\textbf{WSC}} \\
\cmidrule(lr){3-6}\cmidrule(lr){7-10}
& & Base & SAM & GAM & FlatPO & Base & SAM & GAM & FlatPO \\
\midrule
\midrule

\multirow{3}{*}{OPT-1.3B}
& LoRA   & 92.6 & 93.2 & 93.2 & \textbf{94.4} & 62.8 & 63.4 & 63.6 & \textbf{64.4} \\
& Prefix & 92.7 & 92.7 & 93.2 & \textbf{94.0} & 63.4 & 64.0 & 64.4 & \textbf{65.3} \\
& Prompt & 92.6 & 93.0 & 93.1 & \textbf{93.9} & 62.5 & 63.4 & 63.5 & \textbf{64.5} \\
\midrule

\multirow{3}{*}{LLaMA-7B}
& LoRA   & 93.4 & 93.6 & 93.8 & \textbf{94.7} & 65.3 & 65.9 & 65.8 & \textbf{66.5} \\
& Prefix & 92.8 & 93.4 & 93.6 & \textbf{94.8} & 64.4 & 65.2 & 65.5 & \textbf{66.2} \\
& Prompt & 93.2 & 93.6 & 93.8 & \textbf{94.6} & 64.5 & 65.2 & 65.5 & \textbf{65.9} \\
\midrule

\multirow{3}{*}{LLaMA-13B}
& LoRA   & 95.5 & 95.8 & 95.8 & \textbf{96.1} & 65.5 & 67.3 & 68.6 & \textbf{69.2} \\
& Prefix & 94.5 & 94.8 & 95.0 & \textbf{95.4} & 66.0 & 67.3 & 68.4 & \textbf{69.2} \\
& Prompt & 94.3 & 95.1 & 95.1 & \textbf{95.8} & 65.3 & 67.0 & 68.0 & \textbf{68.6} \\

\midrule
\midrule

\multirow{2}{*}{\textbf{MODEL}} & \multirow{2}{*}{\textbf{METHOD}}
& \multicolumn{4}{c}{\textbf{WIC}}
& \multicolumn{4}{c}{\textbf{BoolQ}} \\
\cmidrule(lr){3-6}\cmidrule(lr){7-10}
& & Base & SAM & GAM & FlatPO & Base & SAM & GAM & FlatPO \\
\midrule
\midrule

\multirow{3}{*}{OPT-1.3B}
& LoRA   & 65.0 & 65.5 & 66.2 & \textbf{67.2} & 73.7 & 75.0 & 74.9 & \textbf{75.9} \\
& Prefix & 64.4 & 65.2 & 65.3 & \textbf{66.2} & 73.1 & 74.1 & 74.4 & \textbf{75.2} \\
& Prompt & 63.6 & 64.4 & 64.5 & \textbf{65.3} & 66.2 & 67.5 & 67.5 & \textbf{68.0} \\
\midrule

\multirow{3}{*}{LLaMA-7B}
& LoRA   & 70.5 & 70.8 & 71.0 & \textbf{72.0} & 84.2 & 84.9 & 85.0 & \textbf{86.0} \\
& Prefix & 70.8 & 71.2 & 71.6 & \textbf{72.5} & 83.8 & 84.8 & 84.6 & \textbf{85.4} \\
& Prompt & 69.2 & 69.5 & 69.8 & \textbf{70.5} & 77.3 & 78.6 & 78.8 & \textbf{79.3} \\
\midrule

\multirow{3}{*}{LLaMA-13B}
& LoRA   & 72.5 & 73.0 & 73.3 & \textbf{74.0} & 87.0 & 87.5 & 87.6 & \textbf{88.5} \\
& Prefix & 72.8 & 73.3 & 73.6 & \textbf{74.2} & 86.5 & 87.2 & 87.4 & \textbf{88.2} \\
& Prompt & 71.5 & 72.0 & 72.3 & \textbf{73.0} & 80.5 & 81.2 & 81.7 & \textbf{82.3} \\

\bottomrule[1.2pt]
\end{tabular}
\end{table*}

\begin{table*}[t]
\centering
\setlength{\tabcolsep}{6pt}
\captionsetup{font={scriptsize}, labelsep = newline, justification=centering}
\caption{Fine-tuning performance of \textbf{VLM} with typical PEFT methods (LoRA and Prefix Tuning) on various vision-language benchmarks (ScienceQA, VizWiz, IconQA, Flickr30k, OKVQA, OCRVQA and VQAv2). The best results are highlighted in \textbf{bold}.}
\label{tab:main2}

\newcolumntype{C}[1]{>{\centering\arraybackslash}m{#1}}

\begin{tabular}{C{1.8cm} C{1.6cm} *{8}{C{0.95cm}}}
\toprule[1.2pt]

\multirow{2}{*}{\textbf{MODEL}} & \multirow{2}{*}{\textbf{METHOD}}
& \multicolumn{4}{c}{\textbf{ScienceQA}}
& \multicolumn{4}{c}{\textbf{VizWiz}} \\
\cmidrule(lr){3-6}\cmidrule(lr){7-10}
& & Base & SAM & GAM & FlatPO & Base & SAM & GAM & FlatPO \\
\midrule
\midrule

\multirow{2}{*}{Qwen2.5-VL-3B}
& LoRA   & 90.6 & 91.7 & 91.8 & \textbf{92.4} & 68.7 & 69.8 & 70.6 & \textbf{71.2} \\
& Prefix & 86.2 & 87.8 & 89.8 & \textbf{90.9} & 66.3 & 67.3 & 68.8 & \textbf{70.4} \\
\midrule
\midrule

\multirow{2}{*}{\textbf{MODEL}} & \multirow{2}{*}{\textbf{METHOD}}
& \multicolumn{4}{c}{\textbf{IconQA-txt}}
& \multicolumn{4}{c}{\textbf{IconQA-blank}} \\
\cmidrule(lr){3-6}\cmidrule(lr){7-10}
& & Base & SAM & GAM & FlatPO & Base & SAM & GAM & FlatPO \\
\midrule
\midrule

\multirow{2}{*}{Qwen2.5-VL-3B}
& LoRA   & 91.5 & 92.7 & 92.9 & \textbf{93.7} & 78.9 & 79.7 & 80.8 & \textbf{81.6} \\
& Prefix & 89.5 & 91.0 & 91.2 & \textbf{91.5} & 75.5 & 76.5 & 77.8 & \textbf{78.0} \\
\midrule
\midrule

\multirow{2}{*}{\textbf{MODEL}} & \multirow{2}{*}{\textbf{METHOD}}
& \multicolumn{4}{c}{\textbf{Flickr30k}}
& \multicolumn{4}{c}{\textbf{OKVQA}} \\
\cmidrule(lr){3-6}\cmidrule(lr){7-10}
& & Base & SAM & GAM & FlatPO & Base & SAM & GAM & FlatPO \\
\midrule
\midrule

\multirow{2}{*}{Qwen2.5-VL-3B}
& LoRA   & 101.4 & 104.8 & 105.2 & \textbf{108.6} & 51.1 & 52.2 & 52.0 & \textbf{53.8} \\
& Prefix & 100.5 & 102.5 & 103.6 & \textbf{106.8} & 50.4 & 51.6 & 50.9 & \textbf{52.0} \\
\midrule
\midrule

\multirow{2}{*}{\textbf{MODEL}} & \multirow{2}{*}{\textbf{METHOD}}
& \multicolumn{4}{c}{\textbf{OCRVQA}}
& \multicolumn{4}{c}{\textbf{VQAv2}} \\
\cmidrule(lr){3-6}\cmidrule(lr){7-10}
& & Base & SAM & GAM & FlatPO & Base & SAM & GAM & FlatPO \\
\midrule
\midrule

\multirow{2}{*}{Qwen2.5-VL-3B}
& LoRA   & 68.6 & 70.8 & 70.5 & \textbf{71.4} & 73.6 & 74.4 & 74.5 & \textbf{76.7} \\
& Prefix & 67.2 & 68.2 & 68.3 & \textbf{69.3} & 70.7 & 72.1 & 72.4 & \textbf{73.6} \\

\bottomrule[1.2pt]
\end{tabular}
\end{table*}

\section{Analysis and Discussion}
\label{section5}

\subsection{Implementation of FlatPO}

\textbf{Implementation Details.} We evaluate three LLMs widely used for large-scale tasks, including OPT-1.3B~\cite{zhang2022opt}, LLaMA-7B and LLaMA-13B~\cite{touvron2023llama}. In this work, all experiments are conducted on the GLUE and SuperGLUE benchmarks~\cite{wang2019superglue}. For simplicity, we use SST-2, WSC, BoolQ, WIC to evaluate FlatPO. In addition, to further validate the generality of FlatPO in multimodal scenarios, we extend our experiments to the vision-language setting using Qwen2.5-VL-3B. For the VLM experiments, we group the benchmarks into out-of-domain evaluation datasets and commonly used datasets. The out-of-domain benchmarks including i) ScienceQA~\cite{saikh2022scienceqa}; ii) VizWiz~\cite{gurari2018vizwiz}; iii) IconQA~\cite{lu2021iconqa}, and iv) Flickr30k~\cite{plummer2015flickr30k}. The commonly used benchmarks including i) OKVQA~\cite{marino2019ok}; ii) OCRVQA~\cite{mishra2019ocr}; and iii) VQAv2~\cite{goyal2017making}. Since OCRVQA and VQAv2 are very large, we randomly extract 20k samples from their training sets as the new training sets and another 5k samples from the test sets to create the new test sets. For the LLM experiments, we set the training steps to 20,000, use a batch size of 16, and enable a cosine learning rate scheduler. For the VLM, all experiments were conducted with a global batch size of 128 and each PEFT method is trained for 3 epochs on the fine-tuning dataset. All experiments are conducted on 4 $\times$ NVIDIA A100-40G GPUs.

\textbf{Compared Baselines.} To thoroughly evaluate the effectiveness of FlatPO, we use three typical PEFT methods: LoRA, Prefix Tuning, and Prompt Tuning. Next, we compare the performance of the following optimization algorithms on OPT-1.3B, LLaMA-7B and LLaMA-13B across various PEFTs: i) \textit{Base}: Adam serves as a baseline to compare performance under a standard training paradigm. ii) \textit{SAM}~\cite{foret2020sharpness}: Sharpness-Aware Minimization, which enhances generalization by imposing a smoothing penalty. iii) \textit{GAM}~\cite{zhang2023gradient}: Gradient norm Aware Minimization, an extension of SAM that seeks minima with uniformly small curvature across all directions. iv) \textit{FlatPO}: Our method leverages flatness preference optimization to guide the model convergence process for flat minima.

\textbf{Evaluation Metrics.}
For language tasks, we evaluate the FlatPO using two metrics: Acc. and SPS. Acc. measures the fine-tuned performance on specific tasks. SPS is measured by the number of training samples processed per second, providing insight into the computational overhead of fine-tuning. For vision-language tasks, we follow the standard evaluation protocol of each benchmark: CIDEr for Flickr30K, Accuracy for IconQA-blank, IconQA-txt, OCRVQA, and ScienceQA, and VQA-Score for OKVQA and VizWiz.

\subsection{Main Results}

\begin{figure}[h]
    \centering
    \includegraphics[width=\linewidth]{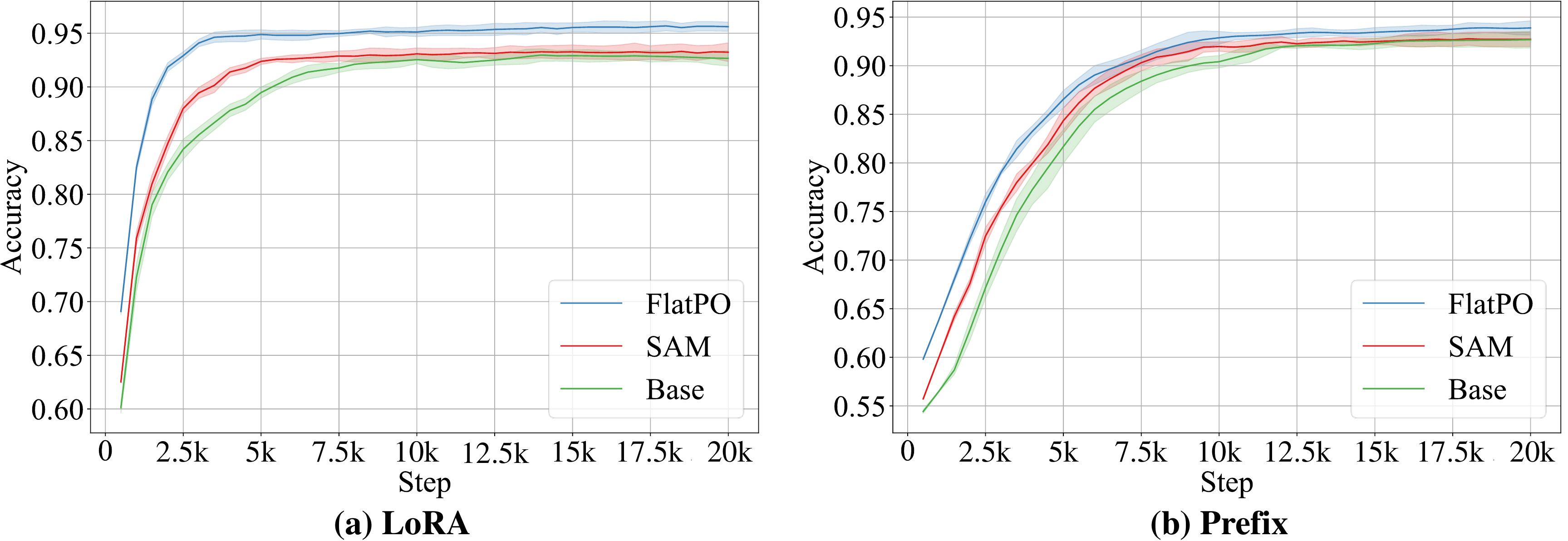}
    \caption{The significance analysis of LoRA and prefix tuning. Perform 3 runs using different random number seeds.}
    \label{fig:significance}
\end{figure}

\textbf{Ability to Capture Flatness Preferences.} We first examine the performance comparison between FlatPO and other methods under various PEFTs. As shown in Table~\ref{tab:main}, FlatPO achieves superior accuracy than baseline optimizers across nearly all experiments, indicating that selectively perturbing only key dimensions can indeed capture the inherent flatness preference in PEFTs. Notably, FlatPO provides stable performance gains across different models and sizes, highlighting its broad applicability. 
Overall, these findings align with our hypothesis that limiting a few critical sharpness values in fine-tuning can lead to tangible performance improvements. 
FlatPO accomplishes this by focusing on the critical subset of parameters that truly drive the sharp directions, thereby promoting convergence toward more generalizable minima. To further assess the versatility of FlatPO beyond pure language tasks, we extend our evaluation to VLM. As shown in Table~\ref{tab:main2}, FlatPO consistently outperforms conventional optimizers across all evaluated datasets. These results underscore that the principle of targeted sharpness reduction is modality-agnostic. FlatPO effectively guides the VLM toward flatter minima without being distracted by the high-dimensional noise inherent in multimodal representations. Consequently, FlatPO emerges as a robust optimization strategy not only for text-based PEFTs but also for adapting large pre-trained vision-language systems. Moreover, we perform 3 runs to evaluate the significance of FlatPO. As shown in Figure~\ref{fig:significance}, FlatPO achieves higher accuracy while exhibiting less variance across multiple runs. These results confirm the superior and stable performance of FlatPO in PEFTs.

\begin{figure}[h]
    \centering
    \includegraphics[width=1.0\linewidth]{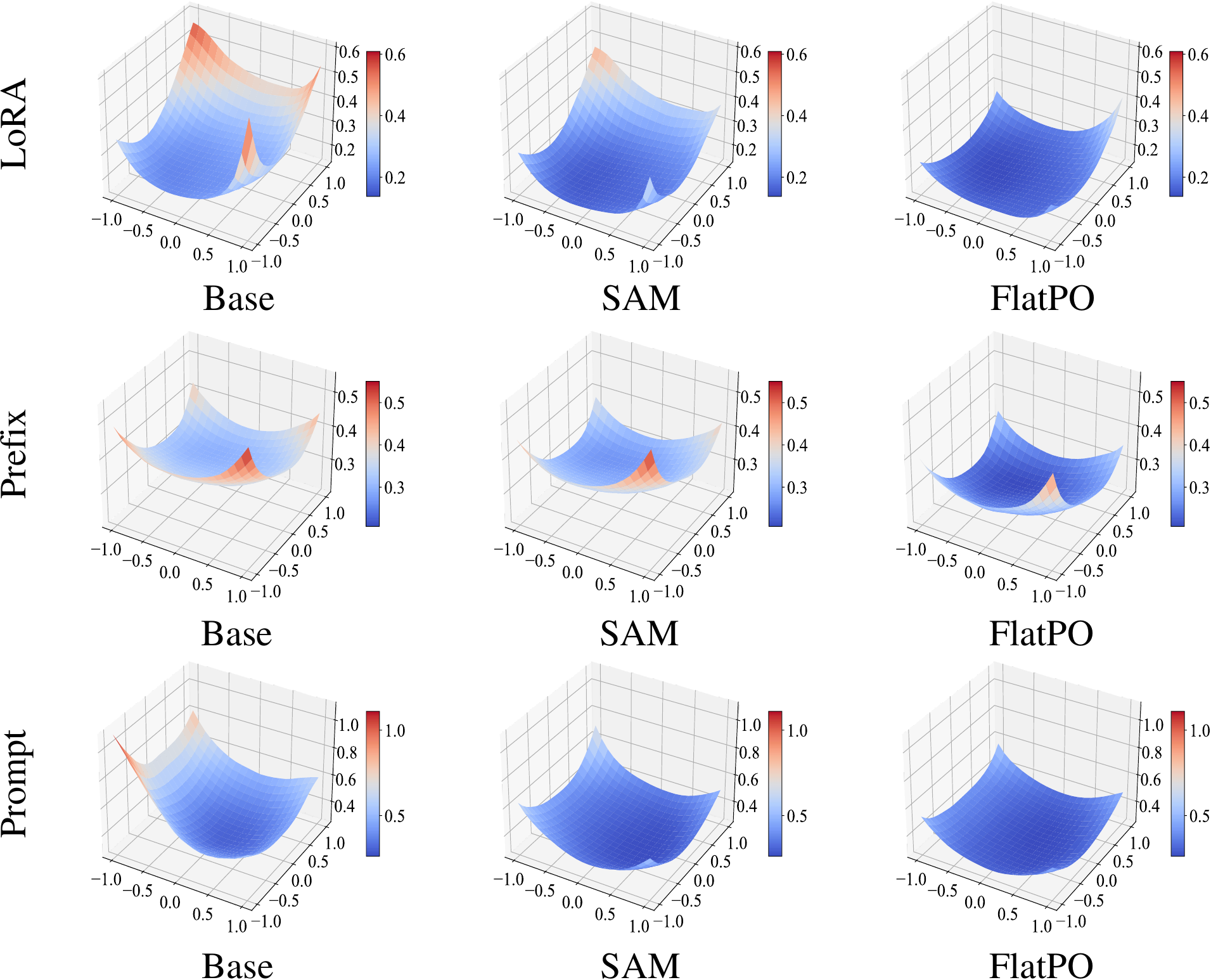}
    \caption{Loss landscapes of LoRA, Prefix Tuning, and Prompt Tuning using Base, SAM, and FlatPO optimization.  }
    \label{fig:losslandscape}
\end{figure}

\textbf{Intuition from Loss Landscape.} For a more intuitive understanding of the optimization process, we visualize the loss landscapes. As shown in Figure~\ref{fig:losslandscape}, our observations indicate that, in PEFT, only a small subset of sensitive dimensions are significantly affected by perturbations, while the majority remain relatively flat. Notably, different PEFTs exhibit unique basin shapes and peak distributions in these low-dimensional subspaces, suggesting variations in their convergence paths and gradient stochasticity. Further comparison reveals that FlatPO tends to converge in flatter regions. This finding implies that intervening in only a few sensitive dimensions can guide the model toward a smoother area of convergence.

\subsection{Ablation Study}

\begin{table}[!h]
\centering
\captionsetup{font={scriptsize}, labelsep = newline, justification=centering}
\caption{Fine-tuning performance with LoRA and Prefix Tuning under different perturbation strategies. The best results are highlighted in \textbf{bold}.}
\setlength{\tabcolsep}{4pt}
\renewcommand{\arraystretch}{1.2}
\begin{tabular}{cccccccc}
\toprule[1.2pt]
METHOD & TYPE & \multicolumn{3}{c}{FlatPO} & \multicolumn{3}{c}{Random} \\ 
\midrule
\midrule
\multicolumn{1}{c}{\multirow{2}{*}{LoRA}} & Weight & Q,V & Q & V & Q,V & Q & V \\
 & Acc. & \textbf{94.4} & 94.0 & 94.2 & \textbf{93.2} & 93.0 & 93.0 \\ 
\hline
\multicolumn{1}{c}{\multirow{2}{*}{Prefix}} & Weight & K,V & K & V & K,V & K & V \\
 & Acc. & \textbf{94.0} & 93.6 & 93.8 & \textbf{92.8} & 92.7 & 92.7 \\ 
\bottomrule[1.2pt]
\end{tabular}
\label{tab:qkv}
\end{table}

\textbf{Comparative Analysis of Targeted vs. Random Perturbation in QKV Components.}
We selectively adjusted specific QKV combinations and compared the effects of targeted perturbation (\textbf{FlatPO}) versus random perturbation (\textbf{Random}). As shown in Table~\ref{tab:qkv}, our experiments demonstrate that: In all settings, FlatPO consistently outperformed Random, indicating that the directionality of the perturbation is crucial for model optimization. Furthermore, adjusting the combinations of components yielded better results than adjusting individual components, suggesting that the interaction relationships among the QKV components are vital, rather than simply the difference in the quantity of the parameters.

\begin{table}[h]
\centering
\setlength{\tabcolsep}{3.5pt}
\renewcommand{\arraystretch}{1.2}
\captionsetup{font={scriptsize}, labelsep = newline, justification=centering}
\caption{Fine-tuning performance with LoRA under different preference ratios. The best results are highlighted in \textbf{bold}.}
\begin{tabular}{ccccccccc}
\toprule[1.2pt]
MODEL                      & TYPE       & \multicolumn{5}{c}{FlatPO}            & SAM   & GAM   \\ 
\midrule
\midrule
\multicolumn{1}{c}{\multirow{3}{*}{OPT-1.3B}}  & ratio      & 1\%   & 5\%   & 10\%  & 20\%  & 60\%  & 100\% & 100\% \\
                           & Acc.   & 93.8        & \textbf{94.4}        & 93.8        & 93.4        & 93.4       & 93.2       & 93.2  \\
                           & SPS. & \textbf{13.06} & 12.97  & 12.80  & 12.21  & 11.56  & 10.62      & 5.35   \\ \hline
\multicolumn{1}{c}{\multirow{3}{*}{LLaMA-7B}}  & ratio      & 1\%   & 5\%   & 10\%  & 20\%  & 60\%  & 100\% & 100\% \\
                           & Acc.   & 94.3        & \textbf{94.7}        & 94.6        & 94.6        & 94.3       & 93.6       & 93.8  \\
                           & SPS. & \textbf{3.22 }  & 3.21   & 3.10   & 2.83   & 2.66  & 2.41       & 1.33   \\ \hline
\multicolumn{1}{c}{\multirow{3}{*}{LLaMA-13B}} & ratio      & 1\%   & 5\%   & 10\%  & 20\%  & 60\%  & 100\% & 100\% \\
                           & Acc.   & 95.8        & \textbf{96.1}        & 96.1        & 95.8        & 95.8       & 95.8       & 95.8  \\
                           & SPS. & \textbf{1.59 } & 1.52   & 1.50   & 1.42   & 1.38  & 1.22       & 0.65   \\                       
\bottomrule[1.2pt]
\end{tabular}
\label{tab:rate}
\vspace{-5pt}
\end{table}

\textbf{Preference Rate Analysis.} 
We ablate perturbation ratios to investigate the effect of different proportions of perturbations on the flatness preference of the model. In detail, we select the top \(\{1\%, 5\%, 10\%, 20\%, 60\%, 100\%\}\) of the gradient parameters for perturbation. As shown in Table~\ref{tab:rate}, perturbing the top 5\% of parameters achieves the best balance between model accuracy and training cost. Although a smaller ratio (\emph{e.g.}, 1\%) is more efficient, slightly compromise performance. Meanwhile, a larger ratio (\emph{e.g.}, 20\%) yields limited performance gains but incurs increased overhead.

\textbf{Efficiency Analysis.} We analyze the efficiency of FlatPO here. In Table~\ref{tab:rate}, a clear trend is consistent efficiency improvement across all ratios. Next, another trend is the consequent improvement in efficiency as the perturbation ratio decreases. Although perturbing only a small percentage of the dimensions should theoretically yield a more significant speedup, this is not entirely the case. While the pre-trained model does not participate in gradient computation, it is still involved in backpropagation in some PEFTs. As a result, the actual speedup does not perfectly correlate with the reduction in perturbation ratio in this case. Still, FlatPO achieves speedups of around 30\% at the optimal 5\% ratio, significantly less than 100\% case. Additionally, while GAM focus on first-order information and offers some benefits in improving the generalization of PEFT, this is not effective in capturing the flatness preference. In addition, GAM is not a favorable candidate due to its inherently high computational overhead.

\begin{table}[h]
\centering
\captionsetup{font={scriptsize}, labelsep = newline, justification=centering}
\caption{Fine-tuning performance with LoRA under different perturbation radius. The best results are highlighted in \textbf{bold}.}
\setlength{\tabcolsep}{6pt}
\renewcommand{\arraystretch}{1.2}
\centering
\begin{tabular}{cccccc}
\toprule[1.2pt]
MODEL                      & TYPE     & \multicolumn{4}{c}{VALUE} \\ \midrule
\midrule
\multicolumn{1}{c}{\multirow{2}{*}{OPT-1.3B}}  & $\rho$   & 0.01 & 0.05 & 0.1  & 0.2  \\
                           & Acc. & 93.6 & 94.0 & \textbf{94.4} & 93.4 \\ \hline
\multicolumn{1}{c}{\multirow{2}{*}{LLaMA-7B}}  & $\rho$   & 0.01 & 0.05 & 0.1  & 0.2  \\
                           & Acc. & 93.8 & 94.6 & \textbf{94.7} & 93.6    \\ \hline
\multicolumn{1}{c}{\multirow{2}{*}{LLaMA-13B}} & $\rho$   & 0.01 & 0.05 & 0.1  & 0.2  \\
                           & Acc. & 95.8 & 96.0 & \textbf{96.1} & 95.8 \\ 
                           \bottomrule[1.2pt]
\end{tabular}
\label{tab:radius}
\end{table}
 
\textbf{Influence of Perturbation Radius.}
The perturbation radius controls the magnitude of the perturbation applied to the model parameters. To verify its effect, we evaluate the effect of different radius values (\(\rho \in \{0.01, 0.05, 0.1, 0.2\}\)) on FlatPO. As shown in  Table~\ref{tab:radius}, a moderate \(\rho\) (\emph{e.g.}, 0.05 or 0.1) can guide FlatPO to converge toward flatter minima, effectively alleviating overfitting without causing substantial training instability. In contrast, an excessively large \(\rho\) (\emph{e.g.}, 0.2) may lead to optimization instability, as overly strong perturbations obscure effective gradient information. 

\begin{table}[h]
\centering
\captionsetup{font={scriptsize}, labelsep = newline, justification=centering}
\caption{Fine-tuning performance with LoRA across different LoRA ranks. The best results are highlighted in \textbf{bold}.}
\setlength{\tabcolsep}{6pt}
\renewcommand{\arraystretch}{1.2}
\begin{tabular}{cccccc}
\toprule[1.2pt]
MODEL                      & TYPE     & \multicolumn{4}{c}{VALUE} \\ \midrule
\midrule
\multicolumn{1}{c}{\multirow{2}{*}{OPT-1.3B}}  & Rank     & 4    & 8    & 16   & 32   \\
                           & Acc. & 93.6 & 94.4 & \textbf{94.5} & 94.3 \\ \hline
\multicolumn{1}{c}{\multirow{2}{*}{LLaMA-7B}}  & Rank     & 4    & 8    & 16   & 32   \\
                           & Acc. & 93.8 & 94.7 & \textbf{95.0} & 94.7   \\ \hline
\multicolumn{1}{c}{\multirow{2}{*}{LLaMA-13B}} & Rank     & 4    & 8    & 16   & 32   \\
                           & Acc. & 95.4 & 96.1 & \textbf{96.2} & 96.1 \\ 
                           \bottomrule[1.2pt]
\end{tabular}
\label{tab:rank}
\vspace{-5pt}
\end{table}

\textbf{Influence of Hyperparameters from PEFTs.}
We investigate the influence of different PEFT hyperparameters on FlatPO. Taking LoRA as our example, the rank of the trainable matrices directly determines the performance of fine-tuning. Thus, we evaluate the effect of different ranks (\{4, 8, 16, 32\}) on FlatPO. As shown in Table~\ref{tab:rank}, as rank increases, the benefits of preference perturbation become more pronounced. A higher rank provides LoRA with a larger representational space, allowing it to make fuller use of the flatter minima discovered by FlatPO.

\subsection{Discussion}

\begin{figure*}[t]
    \centering
    \includegraphics[width=\linewidth]{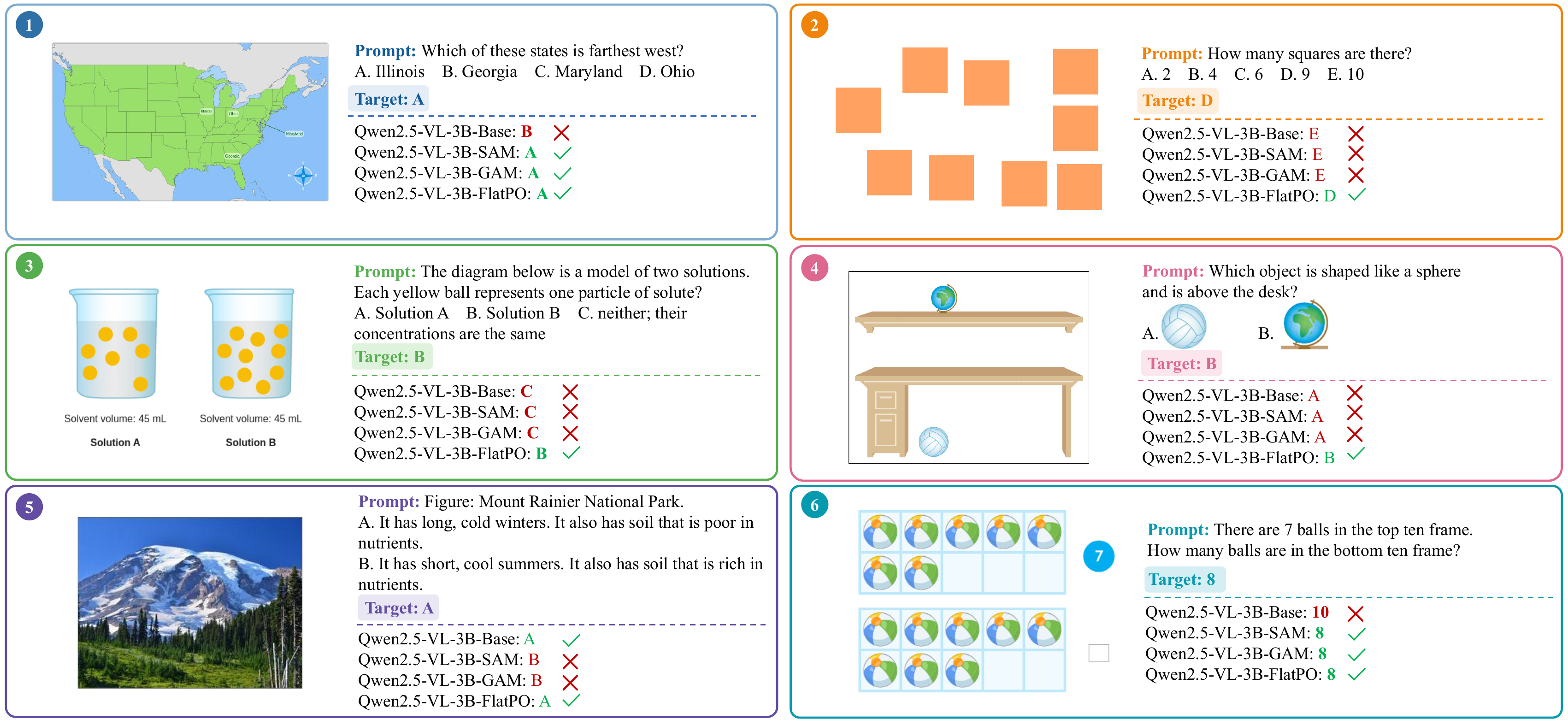}
    \vspace{-10pt}
    \caption{Qualitative comparison on VQA tasks with LoRA using different optimizers: Base, SAM, GAM, and our FlatPO. The examples are selected as representative cases to qualitatively illustrate the differences among these optimization strategies.}
    \label{fig:examples}
    \vspace{-10pt}
\end{figure*}

\begin{figure}[!h]
    \centering
    \includegraphics[width=.8\linewidth]{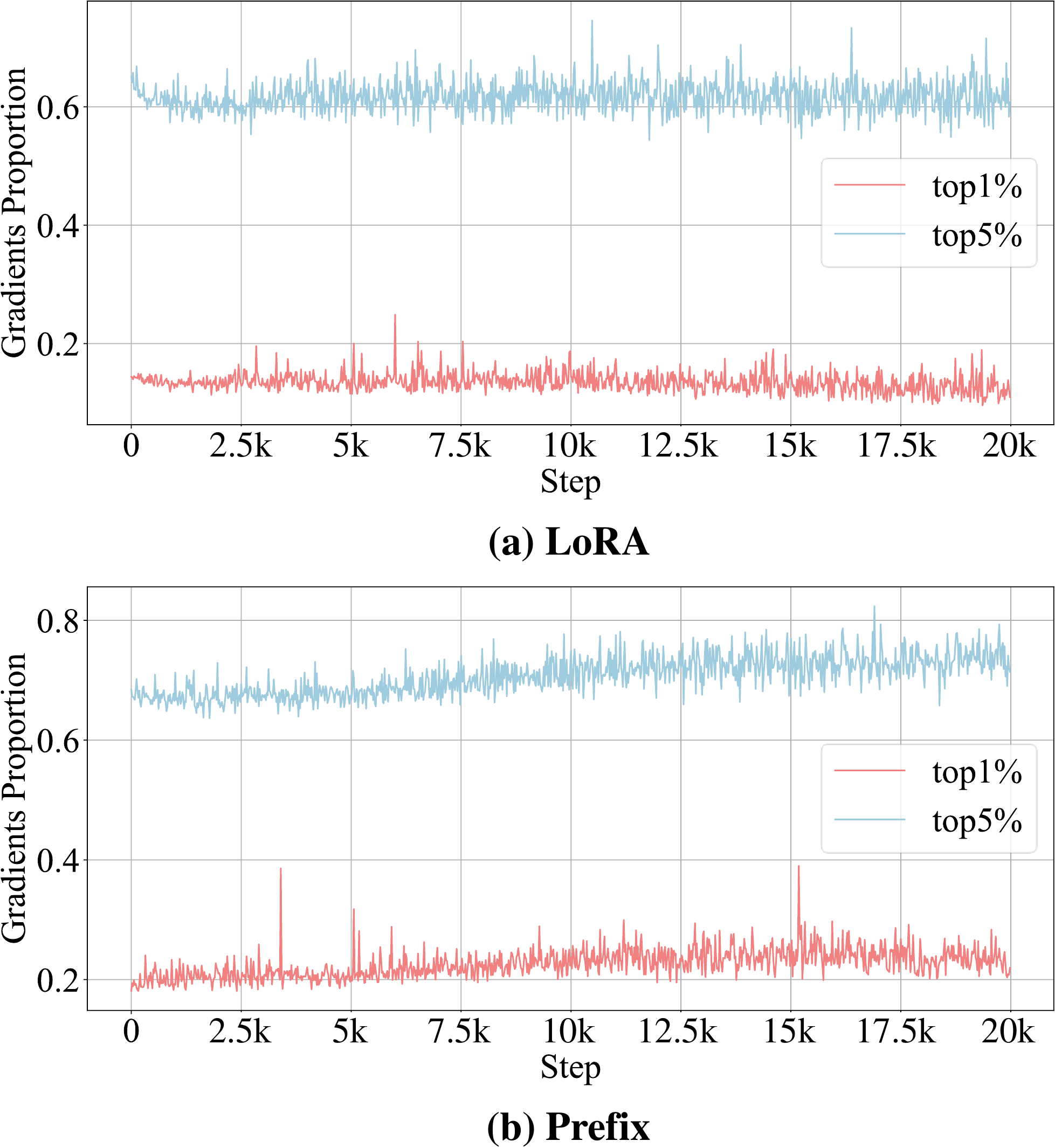}
    \caption{Gradient Contribution of LoRA and Prefix Tuning. At each step, we record the sum of the top 5\% and 1\% gradient norms, then compute their ratios relative to the total gradient norm sum.}
    \label{fig:Gradient}
    \vspace{-5pt}
\end{figure}

\textbf{Further Gradient Analysis.}
To better understand the role of the 5\% perturbation ratio, we further analyzed the gradient contribution during optimization. Figure~\ref{fig:Gradient} illustrates the contribution proportions of the top 5\% and 1\% to the total gradient magnitude. This dynamic analysis reveals two key findings: The top 5\% of parameters consistently contribute over 60\% of the total gradient, demonstrating that PEFT optimization is driven by a sparse subset of critical parameters. The top 1\% parameters show a sharp decline (40\%) in contribution compared to the top 5\%, indicating that excessive sparsity degrades the effective gradient information and 5\% serves as the inflection point for information density.

\textbf{Qualitative Examples on VQA Tasks.}
We present several qualitative examples in Figure~\ref{fig:examples}. The examples are drawn from the evaluated VQA benchmarks. The representative improvements in answer accuracy confirm that FlatPO not only boosts quantitative metrics but also yields more reliable responses in practical VQA scenarios.

\section{Conclusion}
\label{section6}

This paper reveals a prevalent flatness preference in various PEFTs. This finding suggests that focusing on flattening a few sharp dimensions (5\%) is a powerful strategy for improving the generalization of PEFTs rather than attempting to flatten all uniformly (100\%). Based on this, we propose a FlatPO method that guides various PEFTs to converge to a flat area during fine-tuning. Extensive results highlight the practical benefits of FlatPO and underscore the importance of rethinking gradient distribution in PEFTs.

% \section*{Acknowledgments}
% This should be a simple paragraph before the References to thank those individuals and institutions who have supported your work on this article.

\bibliographystyle{IEEEtran}
\bibliography{reference}

\begin{thebibliography}{10}
\providecommand{\url}[1]{#1}
\csname url@samestyle\endcsname
\providecommand{\newblock}{\relax}
\providecommand{\bibinfo}[2]{#2}
\providecommand{\BIBentrySTDinterwordspacing}{\spaceskip=0pt\relax}
\providecommand{\BIBentryALTinterwordstretchfactor}{4}
\providecommand{\BIBentryALTinterwordspacing}{\spaceskip=\fontdimen2\font plus
\BIBentryALTinterwordstretchfactor\fontdimen3\font minus \fontdimen4\font\relax}
\providecommand{\BIBforeignlanguage}[2]{{%
\expandafter\ifx\csname l@#1\endcsname\relax
\typeout{** WARNING: IEEEtran.bst: No hyphenation pattern has been}%
\typeout{** loaded for the language `#1'. Using the pattern for}%
\typeout{** the default language instead.}%
\else
\language=\csname l@#1\endcsname
\fi
#2}}
\providecommand{\BIBdecl}{\relax}
\BIBdecl

\bibitem{han2024parameter}
Z.~Han, C.~Gao, J.~Liu, J.~Zhang, and S.~Q. Zhang, ``Parameter-efficient fine-tuning for large models: A comprehensive survey,'' \emph{arXiv preprint arXiv:2403.14608}, 2024.

\bibitem{wang2025parameter}
L.~Wang, S.~Chen, L.~Jiang, S.~Pan, R.~Cai, S.~Yang, and F.~Yang, ``Parameter-efficient fine-tuning in large language models: a survey of methodologies,'' \emph{Artificial Intelligence Review}, vol.~58, no.~8, p. 227, 2025.

\bibitem{xu2026parameter}
L.~Xu, H.~Xie, S.~J. Qin, X.~Tao, and F.~L. Wang, ``Parameter-efficient fine-tuning methods for pretrained language models: A critical review and assessment,'' \emph{IEEE Transactions on Pattern Analysis and Machine Intelligence}, 2026.

\bibitem{hu2021lora}
E.~J. Hu, Y.~Shen, P.~Wallis, Z.~Allen-Zhu, Y.~Li, S.~Wang, L.~Wang, and W.~Chen, ``Lora: Low-rank adaptation of large language models,'' \emph{arXiv preprint arXiv:2106.09685}, 2021.

\bibitem{lester2021power}
B.~Lester, R.~Al-Rfou, and N.~Constant, ``The power of scale for parameter-efficient prompt tuning,'' \emph{arXiv preprint arXiv:2104.08691}, 2021.

\bibitem{hu2023llm}
Z.~Hu, L.~Wang, Y.~Lan, W.~Xu, E.-P. Lim, L.~Bing, X.~Xu, S.~Poria, and R.~Lee, ``Llm-adapters: An adapter family for parameter-efficient fine-tuning of large language models,'' in \emph{Proceedings of the 2023 conference on empirical methods in natural language processing}, 2023, pp. 5254--5276.

\bibitem{10171397}
Y.~Xing, Q.~Wu, D.~Cheng, S.~Zhang, G.~Liang, P.~Wang, and Y.~Zhang, ``Dual modality prompt tuning for vision-language pre-trained model,'' \emph{IEEE Transactions on Multimedia}, vol.~26, pp. 2056--2068, 2024.

\bibitem{zhang2024unleash}
W.~Zhang, L.~Wu, Z.~Zhang, T.~Yu, C.~Ma, X.~Jin, X.~Yang, and W.~Zeng, ``Unleash the power of vision-language models by visual attention prompt and multimodal interaction,'' \emph{IEEE Transactions on Multimedia}, vol.~27, pp. 2399--2411, 2024.

\bibitem{sun2026unleashing}
C.~Sun, J.~Wei, Y.~Wu, Y.~Shi, S.~He, Z.~Ma, N.~Xie, and Y.~Yang, ``Unleashing the power of singular values for parameter-efficient fine-tuning of large pre-trained models,'' \emph{IEEE Transactions on Multimedia}, 2026.

\bibitem{oikonomou2025sharpness}
D.~Oikonomou and N.~Loizou, ``Sharpness-aware minimization: General analysis and improved rates,'' \emph{arXiv preprint arXiv:2503.02225}, 2025.

\bibitem{liu2025bi}
Y.~Liu, T.~Li, Z.~Huang, Z.~Yang, and X.~Huang, ``Bi-lora: Efficient sharpness-aware minimization for fine-tuning large-scale models,'' \emph{arXiv preprint arXiv:2508.19564}, 2025.

\bibitem{song2023sparse}
W.~Song, Z.~Li, L.~Zhang, H.~Zhao, and B.~Du, ``Sparse is enough in fine-tuning pre-trained large language model,'' \emph{International Conference on Machine Learning}, 2024.

\bibitem{chen2025understanding}
H.~Chen, Y.~Dong, Z.~Wei, Y.~Huang, Y.~Zhang, H.~Su, and J.~Zhu, ``Understanding pre-training and fine-tuning from loss landscape perspectives,'' \emph{arXiv e-prints}, pp. arXiv--2505, 2025.

\bibitem{zhou2024towards}
P.~Zhou, X.~Xie, Z.~Lin, and S.~Yan, ``Towards understanding convergence and generalization of adamw,'' \emph{IEEE Transactions on Pattern Analysis and Machine Intelligence}, 2024.

\bibitem{deng2025eflat}
J.~Deng, Q.~Zhu, J.~Pang, L.~Yang, Z.~Fu, and B.~Zhang, ``Eflat-lora: Efficiently seeking flat minima for better generalization in fine-tuning large language models and beyond,'' \emph{arXiv e-prints}, pp. arXiv--2508, 2025.

\bibitem{li2024flat}
T.~Li, Z.~He, Y.~Li, Y.~Wang, L.~Shang, and X.~Huang, ``Flat-lora: Low-rank adaption over a flat loss landscape,'' \emph{arXiv preprint arXiv:2409.14396}, 2024.

\bibitem{wang2024improving}
M.~Wang, J.~Wang, H.~He, Z.~Wang, G.~Huang, F.~Xiong, Z.~Li, L.~Wu \emph{et~al.}, ``Improving generalization and convergence by enhancing implicit regularization,'' \emph{Advances in Neural Information Processing Systems}, vol.~37, pp. 118\,701--118\,744, 2024.

\bibitem{lee2023domain}
G.~Lee, W.~Jang, J.~H. Kim, J.~Jung, and S.~Kim, ``Domain generalization using large pretrained models with mixture-of-adapters,'' \emph{arXiv preprint arXiv:2310.11031}, 2023.

\bibitem{chen2023ptp}
L.~Chen, H.~Huang, and M.~Cheng, ``Ptp: Boosting stability and performance of prompt tuning with perturbation-based regularizer,'' \emph{arXiv preprint arXiv:2305.02423}, 2023.

\bibitem{saha2026grit}
P.~Saha, C.~Rajbangshi, R.~Goyal, M.~Goyal, A.~Deo, B.~Roy, N.~D. Singh, R.~Goswami, and A.~Das, ``Grit--geometry-aware peft with k-facpreconditioning, fisher-guided reprojection, anddynamic rank adaptation,'' \emph{arXiv preprint arXiv:2601.00231}, 2026.

\bibitem{zhou2025sharpness}
Z.~Zhou, M.~Wang, Y.~Mao, B.~Li, and J.~Yan, ``Sharpness-aware minimization efficiently selects flatter minima late in training,'' in \emph{International Conference on Learning Representations}, vol. 2025, 2025, pp. 20\,949--20\,980.

\bibitem{houlsby2019parameter}
N.~Houlsby, A.~Giurgiu, S.~Jastrzebski, B.~Morrone, Q.~De~Laroussilhe, A.~Gesmundo, M.~Attariyan, and S.~Gelly, ``Parameter-efficient transfer learning for nlp,'' in \emph{International conference on machine learning}.\hskip 1em plus 0.5em minus 0.4em\relax PMLR, 2019, pp. 2790--2799.

\bibitem{wang2022adamix}
Y.~Wang, S.~Mukherjee, X.~Liu, J.~Gao, A.~H. Awadallah, and J.~Gao, ``Adamix: Mixture-of-adapter for parameter-efficient tuning of large language models,'' \emph{arXiv preprint arXiv:2205.12410}, vol.~1, no.~2, p.~4, 2022.

\bibitem{lin2024ctrl}
H.~Lin, J.~Cho, A.~Zala, and M.~Bansal, ``Ctrl-adapter: An efficient and versatile framework for adapting diverse controls to any diffusion model,'' \emph{arXiv preprint arXiv:2404.09967}, 2024.

\bibitem{li2024adapter}
M.~Li, P.~Ye, Y.~Huang, L.~Zhang, T.~Chen, T.~He, J.~Fan, and W.~Ouyang, ``Adapter-x: A novel general parameter-efficient fine-tuning framework for vision,'' \emph{arXiv preprint arXiv:2406.03051}, 2024.

\bibitem{li2021prefix}
X.~L. Li and P.~Liang, ``Prefix-tuning: Optimizing continuous prompts for generation,'' \emph{arXiv preprint arXiv:2101.00190}, 2021.

\bibitem{jia2022visual}
M.~Jia, L.~Tang, B.-C. Chen, C.~Cardie, S.~Belongie, B.~Hariharan, and S.-N. Lim, ``Visual prompt tuning,'' in \emph{European Conference on Computer Vision}.\hskip 1em plus 0.5em minus 0.4em\relax Springer, 2022, pp. 709--727.

\bibitem{CP-Prompt}
Y.~Feng, Z.~Tian, Y.~Zhu, Z.~Han, H.~Luo, G.~Zhang, and M.~Song, ``Cp-prompt: Composition-based cross-modal prompting for domain-incremental continual learning,'' in \emph{Proceedings of the 32nd {ACM} International Conference on Multimedia, {MM} 2024, Melbourne, VIC, Australia, 28 October 2024 - 1 November 2024}.\hskip 1em plus 0.5em minus 0.4em\relax {ACM}, 2024, pp. 2729--2738.

\bibitem{zhang2023adalora}
Q.~Zhang, M.~Chen, A.~Bukharin, N.~Karampatziakis, P.~He, Y.~Cheng, W.~Chen, and T.~Zhao, ``Adalora: Adaptive budget allocation for parameter-efficient fine-tuning,'' \emph{arXiv preprint arXiv:2303.10512}, 2023.

\bibitem{hayou2024lora+}
S.~Hayou, N.~Ghosh, and B.~Yu, ``Lora+: Efficient low rank adaptation of large models,'' \emph{arXiv preprint arXiv:2402.12354}, 2024.

\bibitem{liu2023moelora}
Q.~Liu, X.~Wu, X.~Zhao, Y.~Zhu, D.~Xu, F.~Tian, and Y.~Zheng, ``Moelora: An moe-based parameter efficient fine-tuning method for multi-task medical applications,'' \emph{arXiv preprint arXiv:2310.18339}, 2023.

\bibitem{chen2024llava}
S.~Chen, Z.~Jie, and L.~Ma, ``Llava-mole: Sparse mixture of lora experts for mitigating data conflicts in instruction finetuning mllms,'' \emph{arXiv preprint arXiv:2401.16160}, 2024.

\bibitem{khattak2023self}
M.~U. Khattak, S.~T. Wasim, M.~Naseer, S.~Khan, M.-H. Yang, and F.~S. Khan, ``Self-regulating prompts: Foundational model adaptation without forgetting,'' in \emph{Proceedings of the IEEE/CVF International Conference on Computer Vision}, 2023, pp. 15\,190--15\,200.

\bibitem{roy2023consistency}
S.~Roy and A.~Etemad, ``Consistency-guided prompt learning for vision-language models,'' \emph{arXiv preprint arXiv:2306.01195}, 2023.

\bibitem{bian2024make}
A.~Bian, W.~Li, H.~Yuan, M.~Wang, Z.~Zhao, A.~Lu, P.~Ji, T.~Feng \emph{et~al.}, ``Make continual learning stronger via c-flat,'' \emph{Advances in Neural Information Processing Systems}, vol.~37, pp. 7608--7630, 2024.

\bibitem{ni2024pace}
Y.~Ni, S.~Zhang, and P.~Koniusz, ``Pace: Marrying generalization in parameter-efficient fine-tuning with consistency regularization,'' \emph{Advances in Neural Information Processing Systems}, vol.~37, pp. 61\,238--61\,266, 2024.

\bibitem{peng2025glad}
Y.~Peng, P.~Wang, J.~Liu, and S.~Chen, ``Glad: Generalizable tuning for vision-language models,'' in \emph{Proceedings of the IEEE/CVF International Conference on Computer Vision}, 2025, pp. 4310--4320.

\bibitem{foret2020sharpness}
P.~Foret, A.~Kleiner, H.~Mobahi, and B.~Neyshabur, ``Sharpness-aware minimization for efficiently improving generalization,'' \emph{arXiv preprint arXiv:2010.01412}, 2020.

\bibitem{zhang2023gradient}
X.~Zhang, R.~Xu, H.~Yu, H.~Zou, and P.~Cui, ``Gradient norm aware minimization seeks first-order flatness and improves generalization,'' in \emph{Proceedings of the IEEE/CVF Conference on Computer Vision and Pattern Recognition}, 2023, pp. 20\,247--20\,257.

\bibitem{ding2023parameter}
N.~Ding, Y.~Qin, G.~Yang, F.~Wei, Z.~Yang, Y.~Su, S.~Hu, Y.~Chen, C.-M. Chan, W.~Chen \emph{et~al.}, ``Parameter-efficient fine-tuning of large-scale pre-trained language models,'' \emph{Nature Machine Intelligence}, 2023.

\bibitem{zhang2022opt}
S.~Zhang, S.~Roller, N.~Goyal, M.~Artetxe, M.~Chen, S.~Chen, C.~Dewan, M.~Diab, X.~Li, X.~V. Lin \emph{et~al.}, ``Opt: Open pre-trained transformer language models,'' \emph{arXiv preprint arXiv:2205.01068}, 2022.

\bibitem{touvron2023llama}
H.~Touvron, T.~Lavril, G.~Izacard, X.~Martinet, M.-A. Lachaux, T.~Lacroix, B.~Rozi{\`e}re, N.~Goyal, E.~Hambro, F.~Azhar \emph{et~al.}, ``Llama: open and efficient foundation language models. arxiv,'' \emph{arXiv preprint arXiv:2302.13971}, 2023.

\bibitem{wang2019superglue}
A.~Wang, Y.~Pruksachatkun, N.~Nangia, A.~Singh, J.~Michael, F.~Hill, O.~Levy, and S.~Bowman, ``Superglue: A stickier benchmark for general-purpose language understanding systems,'' \emph{Advances in neural information processing systems}, vol.~32, 2019.

\bibitem{saikh2022scienceqa}
T.~Saikh, T.~Ghosal, A.~Mittal, A.~Ekbal, and P.~Bhattacharyya, ``Scienceqa: A novel resource for question answering on scholarly articles,'' \emph{International Journal on Digital Libraries}, vol.~23, no.~3, pp. 289--301, 2022.

\bibitem{gurari2018vizwiz}
D.~Gurari, Q.~Li, A.~J. Stangl, A.~Guo, C.~Lin, K.~Grauman, J.~Luo, and J.~P. Bigham, ``Vizwiz grand challenge: Answering visual questions from blind people,'' in \emph{Proceedings of the IEEE conference on computer vision and pattern recognition}, 2018, pp. 3608--3617.

\bibitem{lu2021iconqa}
P.~Lu, L.~Qiu, J.~Chen, T.~Xia, Y.~Zhao, W.~Zhang, Z.~Yu, X.~Liang, and S.-C. Zhu, ``Iconqa: A new benchmark for abstract diagram understanding and visual language reasoning,'' \emph{arXiv preprint arXiv:2110.13214}, 2021.

\bibitem{plummer2015flickr30k}
B.~A. Plummer, L.~Wang, C.~M. Cervantes, J.~C. Caicedo, J.~Hockenmaier, and S.~Lazebnik, ``Flickr30k entities: Collecting region-to-phrase correspondences for richer image-to-sentence models,'' in \emph{Proceedings of the IEEE international conference on computer vision}, 2015, pp. 2641--2649.

\bibitem{marino2019ok}
K.~Marino, M.~Rastegari, A.~Farhadi, and R.~Mottaghi, ``Ok-vqa: A visual question answering benchmark requiring external knowledge,'' in \emph{Proceedings of the IEEE/cvf conference on computer vision and pattern recognition}, 2019, pp. 3195--3204.

\bibitem{mishra2019ocr}
A.~Mishra, S.~Shekhar, A.~K. Singh, and A.~Chakraborty, ``Ocr-vqa: Visual question answering by reading text in images,'' in \emph{2019 international conference on document analysis and recognition (ICDAR)}.\hskip 1em plus 0.5em minus 0.4em\relax IEEE, 2019, pp. 947--952.

\bibitem{goyal2017making}
Y.~Goyal, T.~Khot, D.~Summers-Stay, D.~Batra, and D.~Parikh, ``Making the v in vqa matter: Elevating the role of image understanding in visual question answering,'' in \emph{Proceedings of the IEEE conference on computer vision and pattern recognition}, 2017, pp. 6904--6913.

\end{thebibliography}

\newpage

 

% \end{IEEEbiography}



\vfill

\end{document}